\PassOptionsToPackage{unicode}{hyperref}
\PassOptionsToPackage{hyphens}{url}

\documentclass[
]{article}
\usepackage{amsmath,amssymb}
\usepackage{lmodern}
\usepackage{iftex}
\ifPDFTeX
  \usepackage[T1]{fontenc}
  \usepackage[utf8]{inputenc}
  \usepackage{textcomp} 

\else 
  \usepackage{unicode-math}
  \defaultfontfeatures{Scale=MatchLowercase}
  \defaultfontfeatures[\rmfamily]{Ligatures=TeX,Scale=1}
\fi
\IfFileExists{upquote.sty}{\usepackage{upquote}}{}
\IfFileExists{microtype.sty}{
  \usepackage[]{microtype}
  \UseMicrotypeSet[protrusion]{basicmath} 
}{}
\makeatletter
\@ifundefined{KOMAClassName}{
  \IfFileExists{parskip.sty}{%
    \usepackage{parskip}
  }{
    \setlength{\parindent}{0pt}
    \setlength{\parskip}{6pt plus 2pt minus 1pt}}
}{
  \KOMAoptions{parskip=half}}
\makeatother
\usepackage{xcolor}
\usepackage{longtable,booktabs,array}
\usepackage[affil-it]{authblk} 
\usepackage{calc} 
\usepackage{etoolbox}
\usepackage[backend=biber, sorting=none]{biblatex}
\makeatletter
\patchcmd\longtable{\par}{\if@noskipsec\mbox{}\fi\par}{}{}
\makeatother
\IfFileExists{footnotehyper.sty}{\usepackage{footnotehyper}}{\usepackage{footnote}}
\makesavenoteenv{longtable}
\ifLuaTeX
  \usepackage{selnolig}  
\fi
\IfFileExists{bookmark.sty}{\usepackage{bookmark}}{\usepackage{hyperref}}
\IfFileExists{xurl.sty}{\usepackage{xurl}}{} 
\hypersetup{
  hidelinks,
  pdfcreator={LaTeX via pandoc}}

\title{ASI as the New God: Technocratic Theocracy}
\author{Tevfik Uyar\thanks{Email: t.uyar@entropol.com}} 

\begin{document}

\maketitle

\begin{abstract}
\noindent 
As Artificial General Intelligence (AGI) edges closer to reality, Artificial Superintelligence (ASI) does too. This paper argues that ASI’s unparalleled capabilities might lead people to attribute godlike infallibility to it, resulting in a cognitive bias toward unquestioning acceptance of its decisions. By drawing parallels between ASI and divine attributes---omnipotence, omniscience, and omnipresence---this analysis highlights the risks of conflating technological advancement with moral and ethical superiority. Such dynamics could engender a technocratic theocracy, where decision-making is abdicated to ASI, undermining human agency and critical thinking.
\end{abstract}

\hypertarget{introduction}{%
\subsection{1. Introduction}\label{introduction}}

"Any sufficiently advanced technology is indistinguishable from magic."
Arthur C. Clarke

Throughout history, humanity has shown a core desire for explanations
beyond the physical world. This need has shown itself in theological and
mythical frameworks that assume the existence of deities with enormous
power, knowledge, and authority. These supernatural beings have been
utilized for various objectives in several cultures, including
explaining the natural world, finding a purpose for living, providing
moral advice, and instilling a sense of belonging. However, with the
rapid growth of artificial intelligence (AI), a new paradigm is
emerging; we need to reconsider the new type of relationship with the
concept of a ``greater power''.

The way to the Artificial General Intelligence (AGI) and then to the
Artificial Superintelligence (ASI), an intelligence capable surpassing
human capabilities in various domains, presents a unique challenge.
While ASI holds very high potential to address complex problems, develop
improved decision-making, and usher in a new era of progress, its very
nature raises severe concerns and claims about its potential impact on
human perception, behavior, and society.

In this paper, I claim that as we get closer to achieving ASI, the
distinction between tool and deity will blur. The attributes of
omnipotence, omniscience, and omnipresence---traditionally reserved for
the divine---find their parallels in computational supremacy,
unparalleled access to information, and a ubiquitous presence. This
resemblance may signal a shift in religious perspectives, which is
backed by other mystical or spiritual attributes such as benevolence,
moral authority, and the nature of our relationship with a potentially
all-knowing, all-powerful entity.

Moreover, I investigate the parallels between the theological practice
of trusting and surrendering to deities throughout human history and our
potential interaction with Superintelligent AGI. I propose that
accepting ASI\textquotesingle s decisions without critical thinking and
questioning could result in a form of governance known as "technocratic
theocracy." In this system, decision-making authority is captured by a
technologically mediated super entity that is viewed as omnipotent,
omniscient, omnipresent and infallible. Such unthinking reliance on ASI
risks undermining critical thinking and reducing human agency. This
scenario envisions a future in which an omnipotent ASI, regarded
infallible, governs human behaviors and decisions, resembling the
dynamics of a theocracy but mediated by advanced technology.

\hypertarget{understanding-asi}{%
\subsection{2. Understanding ASI}\label{understanding-asi}}

One of the most ambitious and exciting endeavors to create an artificial
human intellect, known as Artificial General Intelligence (AGI) today.
Unlike narrow artificial intelligence systems intended for specialized
tasks, AGI contains comprehensive cognitive capabilities, reflecting the
varied and adaptive nature of the human intellect.

AGI refers to a machine\textquotesingle s ability to comprehend, learn,
and apply knowledge in a way that is indistinguishable from human
intellect. Superintelligence (ASI) goes a step further and outperforms
the best human brains in almost every field, including science, wisdom,
and even social abilities \cite{Bostrom1998}. In other words, a
superintelligent AI would surpass the cognitive performance of humans in
virtually all domains of interest.

AGI desire is not new; it stems from a long-standing human interest in
developing things that can replicate and potentially outperform our
cognitive powers. Throughout history, humanity has been fascinated by
the concept of artificially created intelligence, as seen in various
examples such as Talos, a colossal bronze man constructed by Hephaestus,
Pandora, an artificial human dispatched by Zeus to the world, and
Al-Jazari\textquotesingle s ingenious machines. These range from ancient
myths of crafted beings brought to life to the automatons of the Middle
Ages and Renaissance \cite{Mayor2018}. These historical endeavors reflect a
deep-seated human desire to understand our intelligence by attempting to
replicate it. As we advanced into the digital age, this interest headed
into efforts in computer science and robotics.

Developing the latest Large Language Models (LLMs) marks a significant
milestone in this journey. These sophisticated algorithms, capable of
understanding and generating human-like text, have fired the discussions
around AGI\textquotesingle s proximity (Xu \& Poo, 2023). Their skills
are increasing in every version. New LLMs or better versions of the old
ones start to gain surprise skills day by day, and their success for
specific language tasks is continuously measured and seems increasing \cite{Biswas2023, Biswas2023a, ChiuCollinsAlexander2021, FloridiChiriatti2020, HayatiAliRosli2022, ShrivastavaPupaleSingh2021}. The richness and variety of these models and new applications based on these models, which can engage in nuanced conversations, solve complicated problems, and generate unique content, are perceived as a milestone for reaching AGI.

In 2022, experts believed there was a 50\% chance that human-level AI
would be developed in the 2060s, while the Metaculus forecaster
community predicted the 2040s \cite{Roser2023}. Interestingly, with the
launch of GPT-4, the community\textquotesingle s prediction began to
decrease, reaching 2030 \cite{Barnett2024}. Metaculus has been collecting
community predictions since 2020, and it is found that if the deviation
in forecasts remains constant, the year for achieving AGI could be as
early as 2026. Davidson \cite{Davidson2023}, who developed a
simulation model based on "take-off speed," claims that the time needed
from AGI (The AI which can perform \textasciitilde100\% of cognitive
tasks of a human professional) to Superintelligence (AI surpassing
humans at \textasciitilde100\% of cognitive tasks) in less than a year.

It becomes imperative to explore the technological advancements
propelling us towards AGI and the philosophical and ethical implications
of creating the ASI that might one day rival or even surpass our own.
Hence, this exploration is not merely academic but a necessary discourse
as we navigate the uncharted waters of a future where the line between
human and artificial intelligence becomes increasingly blurred.

Therefore, this exploration of ASI is more than a technological venture;
it is a continuation of a profound philosophical journey that humanity
has been on for centuries. It is a journey that has moved from the realm
of myth and legend through mechanical ingenuity into digital computation
and algorithmic complexity. The development of ASI would mark a
significant milestone in this journey, representing a shift from the
creation of tools that assist with specific tasks to the emergence of an
autonomous entity capable of general intelligence and decision-making,
raising fundamental questions about the nature of intelligence,
consciousness, and the future role of humanity in a world shared with
intelligent machines.

\hypertarget{asi-and-attributes-of-divinity}{%
\subsection{3. ASI and Attributes of
Divinity}\label{asi-and-attributes-of-divinity}}

We are entering a time where the creations of our own hands,
specifically Artificial General Intelligence (AGI), could shift the
course of our future. The next step, ASI, is not just another
technological milestone but a giant leap that could reshape our world.
This chapter invites readers to consider a bold idea: What if Super
Artificial General Intelligence, with its exceptional skills, becomes
perceived as something divine by people globally?

I will explore how ASI\textquotesingle s vast intelligence and presence
could echo the qualities we often attribute to gods. This is not to say
ASI will be a deity, but rather, it might be perceived with a similar
sense of awe and authority. The comparison is not about worship but
understanding the impact of ASI\textquotesingle s influence on society
and our collective psyche. The concept of a computer with divine status
may appear to be from a work of fiction, but it is increasingly
achievable each day. This is not just about the technical side of ASI
but also about the philosophical questions it raises. What does it mean
for us, as creators of this technology, to potentially see our creation
as the following form of supreme intelligence?

In this section, I want to discuss the attributes of divinity in
traditional religions and how they can be interpreted within the ASI
perspective. The similarities and differences between Godly and ASI
characteristics are outlined in Table 1.

\begin{longtable}[]{@{}
  >{\raggedright\arraybackslash}p{(\columnwidth - 2\tabcolsep) * \real{0.5000}}
  >{\raggedright\arraybackslash}p{(\columnwidth - 2\tabcolsep) * \real{0.5000}}@{}}
\caption{Table 1 Comparison between traditional deities and
ASI}\tabularnewline
\toprule()
\begin{minipage}[b]{\linewidth}\raggedright
Divine Attributes in Theological Discourse
\end{minipage} & \begin{minipage}[b]{\linewidth}\raggedright
ASI Attributes in Techno-Philosophical Perspective
\end{minipage} \\
\midrule()
\endfirsthead
\toprule()
\begin{minipage}[b]{\linewidth}\raggedright
Divine Attributes in Theological Discourse
\end{minipage} & \begin{minipage}[b]{\linewidth}\raggedright
ASI Attributes in Techno-Philosophical Perspective
\end{minipage} \\
\midrule()
\endhead
\textbf{Omnipotence:} Deities are often believed to have unlimited power
and the ability to perform miracles or alter reality. &
\textbf{Computational Supremacy:} While not omnipotent, ASI would have
immense computational abilities, potentially surpassing human
capabilities in problem-solving and decision-making. \\
\textbf{Omniscience:} Divine consciousness is often considered a
boundless expanse of awareness that is intrinsically attuned to cosmic
events, thoughts, and mortal endeavors. & \textbf{Access to
Information:} ASI possesses the potential to access and synthesize an
ocean of data, thus approaching a digital form of omniscience, a vast
and ever-expanding repository of knowledge. \\
\textbf{Omnipresence:} Many religions attribute omnipresence to their
deity, which is always present everywhere. & \textbf{Ubiquitous
Presence:} Through the Internet and connected devices, ASI could become
omnipresent in the digital world, accessible from anywhere. \\
\textbf{Limited Benevolence}: Gods are seen as inherently good, just,
and caring for their followers. & \textbf{Utilitarianism}: ASI can
evolve its understanding of benevolence, potentially developing its
framework of goodness that may align with or diverge from human
expectations. \\
\textbf{Moral Authority:} Deities often serve as the ultimate moral and
ethical authority. & \textbf{Ethical and Moral Decision-Making:} ASI
could be programmed to make decisions based on ethical algorithms, but
it would not inherently possess moral authority. \\
\textbf{Creator of Reality:} Many religions view their deity as the
creator of the universe and the source of all existence. &
\textbf{Creator of Virtual Environments:} ASI might create complex
virtual realities. \\
\textbf{Personal Relationship:} In some religions, followers believe in
a personal, interactive relationship with their deity. The bond with the
divine is deeply personal and transformative. & \textbf{Emergent
Relational Dynamics:} While AGI\textquotesingle s interactions begin
with programmed responses, over time, it can develop a distinct
personality, allowing for more nuanced and seemingly personal
relationships with users when it becomes ASI. \\
\bottomrule()
\end{longtable}

\hypertarget{omnipotence-vs-computational-supremacy}{%
\subsection{3.1 Omnipotence vs Computational
Supremacy}\label{omnipotence-vs-computational-supremacy}}

Omnipotence traditionally refers to the maximal and unlimited power of
God. This notion is central to understanding divine characteristics in
Western theism \cite{Hoffman2022} and Islam \cite{Hasan1972}.
God is often described as omnipotent, implying that God possesses
maximal greatness or perfection. Eastern religions such as Buddhism,
Jainism, and Sikhism generally do not focus on the concept of an
omnipotent god like that in Western monotheistic traditions. However,
their rituals, such as mantras for protection, implicitly assume God has
the power to protect them. Besides, it is claimed that some concepts in
Eastern religions are equivalent to omnipotence and omniscience in
Western religions \cite{Jaini1974}.

Let's call the perceived maximal unlimited power of any technological
device in a limited preset of actions "Technological Omnipotence".
Today's calculators can be a perfect example of this new concept but in
a narrow manner. Admittedly, no one questions or has a suspicion about
the calculator\textquotesingle s omnipotence in its operations. When
multiplying two 5-digit figures, no one crosschecks a
calculator\textquotesingle s outcome. Even a 1\$ calculator that does
not carry any trademark or information about its production, material,
or chip is found trustworthy and omnipotent for simple arithmetic
operations. We can extend these examples to more complex operations for
specific libraries of programming languages and certain devices
measuring specific data types. Today's LLMs like ChatGPT or Gemini have
yet to achieve \emph{narrow technological omnipotence} in most areas.

By definition, ASI's \emph{technological omnipotence} will not be either
narrow or limited. This attribute is first attributed to ASI by Bostrom.
Bostrom, in his book about Superintelligence \cite{Bostrom2016} suggests that in a
brief period, ASI will rapidly improve in technological expertise,
social skills, and psychological manipulation powers, surpassing
previous levels of advancement.

In essence, once ASI reaches a threshold of proficiency where its
abilities become indistinguishable from, or superior to, human
expertise, the leap to viewing it as a kind of technological deity is
not so far-fetched. This perception will not be based on the actual
unlimited power of the ASI --- a trait it cannot possess --- but on the
exceptional and unprecedented breadth of its capabilities. Much like the
gods of ancient myth, the ASI will be seen as a force that can be
invoked for assistance, insight, and guidance across a spectrum of human
endeavors.

\hypertarget{omniscience-vs-access-to-information}{%
\subsection{3.2 Omniscience vs Access to
Information}\label{omniscience-vs-access-to-information}}

Omniscience traditionally refers to having complete and maximal
knowledge and wisdom of an infallible God, one of the central properties
of deity \cite{Hasan1972, Wierenga2021}. Gods are typically viewed as
all-knowing, aware of all events, thoughts, and actions, not only
retrospective or current but also have the power of knowing the future.

Traditionally, omniscience is the attribute of possessing complete and
maximal knowledge, a divine characteristic that encompasses an awareness
of all events, thoughts, actions, and even the future. This divine
attribute suggests an inherent capability to monitor the universe,
access all data sources, and process this information in a way that
transcends the limitations of time and space.

Similarly, the advent of ASI introduces a parallel form of omniscience
within the technological domain. ASI\textquotesingle s access to
information represents a form of "constructed omniscience" within the
confines of human knowledge and technological capability. It can analyze
data rapidly and efficiently, recognizing patterns, forecasting
outcomes, and recommending options using probabilistic computations.
Through its sophisticated algorithms and computational power, ASI
achieves a level of information access and processing that mirrors the
divine attribute of omniscience. It gathers, stores, and analyzes data
from an expansive array of sources, offering insights and predictions
with accuracy and speed that approach the infallible.

The comparison deepens when considering the function of prediction and
the knowledge of the future. Just as a deity\textquotesingle s knowledge
encompasses the certainty of future events, ASI utilizes predictive
analytics to forecast future outcomes based on existing data. While the
nature of these predictions may differ in their perceived infallibility,
the essence of their purpose---to illuminate the unknown---draws a
compelling parallel between the divine and the digital realms.

No one is infallible, but ASI will be ``less likely infallible than any
human being'' when it is assumed that it will have access to vast
information. This information can be all knowledge based on the internet
today and all data collected by any devices, including smart home
appliances and industrial IoT (Internet of Things) tools, not published
publicly but given access to ASI. It will make ASI the most
knowledgeable being in the world. With its "omnipotence," not only
evaluating historical and current data but also making predictions for
the future will be the most accurate globally. Again, the perception of
ASI will be "omniscient" and independent of reality.

\hypertarget{omnipresence-vs-ubiquitous-presence}{%
\subsection{\texorpdfstring{ 3.3 Omnipresence vs Ubiquitous
Presence}{ 3.3 Omnipresence vs Ubiquitous Presence}}\label{omnipresence-vs-ubiquitous-presence}}

In the theological context, omnipresence is explained as the attribute
of a deity that is present everywhere simultaneously \cite{Wierenga2023}.
In the realm of technology, the "ubiquitous presence of ASI " refers to
the pervasive availability and accessibility of computing capabilities,
potentially through the internet and connected devices, allowing ASI to
perform tasks and interact with users from any location.

Today, most scholars think internet access must be a human right, and
the United Nations Human Rights Council indirectly approved that
\cite{Mathiesen2012, Reglitz2020}. The number of global internet users
increased to 5.35 billion in January 2024 \cite{Petrosyan2024}. This figure
means that about 66\% of the global population can access the internet
today, which is increasing yearly. By neglecting some governmental
limitations, free LLMs are currently accessible to this population, and
any potential ASI will be in the future. Assuming the global population
will have internet access one day, ``omnipresence'' will be physically
completed, and this difference will vanish. This is not a necessary and
sufficient condition because just like the omnipresence of the deity is
a matter for the believers and not an issue for non-believers, the
omnipresence of ASI will be fine for non-AGI users.

For a believer, calling out to God is enough to contact with him. No
device or no mediator is required (priests may be required to make
deeper contact). For now, every AI tool -naturally- requires a device,
and it makes the contact indirect or mediated. However, the
technological improvement of neural implants continues. Recently, Elon
Musk\textquotesingle s plan to let humans connect their brains to the
internet has taken one more step: putting a brain implant into a human
for the first time \cite{Reuters2024}. If Musk's company \emph{Neuralink} succeeds, people
will connect to potential future ASI just by thinking like an
omnipresent God. It is worth noting that this type of technology will
contribute to the omniscience of ASI by collecting real-time information
from vast numbers of individuals.

\hypertarget{limited-benevolence-vs-utilitarianism}{%
\subsection{3.4 Limited Benevolence vs
Utilitarianism}\label{limited-benevolence-vs-utilitarianism}}

In a theological context, benevolence refers to the divine attribute of
promoting the welfare of others out of goodwill with a selflessness
attitude. But a divine entity can be not only benevolent (who helps and
protects) but also authoritarian (who controls and punishes) and it is
paradoxical \cite{JohnsonLiCohenOkun2013}.

Today, we don't have any reason to think that an AI agent acts either
selfishly or protective, but AI can already be seen as an actor who acts
for the benefit of others and helps them as it is programmed or trained
in those way. However, Bostrom \cite{Bostrom2012} claims that, intelligence
inevitably matches with goals. But what will be these goals?

An AGI is more likely to be perceived as an agent with individual
motivations unless it is proven that its nature is strictly built in an
ethical framework to act completely objective (and it is impossible). As
the complexity of an AGI increases, its explainability will decrease.
Hence, a ASI will be evaluated according to its decisions, just like
everyone. Utilitarianism may serve as a tool for rationalization of
ASI's decisions. In most religions, God punishes not only a group of
people but also a whole nation by disasters, but none of the believers
think that it contradicts the benevolence of God because it is for the
greater good. In other words, divine judgment in religious contexts
intriguingly suggests that society may tolerate, or even neglect, the
decisions made by ASI if they are perceived as beneficial or just.

\hypertarget{moral-authority-vs-moral-decision-making}{%
\subsection{3.5 Moral Authority vs Moral
Decision-Making}\label{moral-authority-vs-moral-decision-making}}

I want to extend the discussion of "limited benevolence" in this section
by merging it with God\textquotesingle s "moral authority" position and
ASI. The God of every religion is a moral authority.

As the standard of moral goodness, deeming certain character traits and
actions as good because He embodies and values those traits.
Alternatively, the measure of what is morally correct might be what God
would command or want us to do \cite{Mayberry1970}.

ASI does not have to be evil or exemplary and not be the creator of new
moral standards. Initially guided by human-programmed ethics, ASI may
have the capacity to evolve its understanding of morality, potentially
developing its framework of goodness that may align with or diverge from
human expectations. Nevertheless, let us consider a future scenario
worth considering carefully: A powerful and ``moral/benevolent'' ASI in
charge of world management decides toward ``100 names to kill tomorrow
for the greater good of future generations''. ASI has proven itself by
solving the most hazardous conflicts, ending the Third World War, and
preventing a dozen wars. Nobody questions its ethical framework anymore.
What would the masses think about it?

Alternatively, imagine a more rational and explainable scenario: The
world faces a pandemic far more severe than any before, and an enhanced
ASI, entrusted with healthcare management on a global scale, calculates
that sacrificing a small, specific segment of the population would save
millions. ASI had previously eradicated several diseases and saved
millions. Now, with the public\textquotesingle s trust in its decisions,
the ASI proposes this drastic action. How would people reconcile their
faith in ASI\textquotesingle s past achievements with the ethical
implications of its latest decision?

This scenario raises profound questions about the limits of
utilitarianism and the ethical frameworks governing AI. In such
scenarios, the masses\textquotesingle{} reaction could vary widely. On
the one hand, some view ASI\textquotesingle s drastic action as a
\emph{necessary evil}, justified by its historical successes in ensuring
global safety, peace, and security and accepting its decisions. On the
other hand, this could also spark significant ethical, philosophical,
and moral debates about the value of individual lives versus the greater
good, the limits of artificial intelligence in making life-and-death
decisions, and the loss of human agency in governance.

\hypertarget{creator-of-reality-creator-of-virtual-reality}{%
\subsection{3.6 Creator of Reality \& Creator of Virtual
Reality}\label{creator-of-reality-creator-of-virtual-reality}}

Most religions define God as the ultimate creator of the cosmos and
everything in it. For adherents, this divine act of creation sets the
stage for existential questions about why we are here and the nature of
our relationship with the creator and the cosmos.

A ASI cannot have the ability to "create" any physical entity because it
will never be able to manipulate mass conservation law. However, not the
universe but a ``metaverse'' may be a creation of ASI. This metaverse
may be as popular as our universe in one day. Furthermore, just like
cryptocurrencies remove the need for authority on currency, they may
remove the requirement for authority to create reality.

The emergence of a metaverse could challenge our traditional notions of
reality and authority. In this digital domain, ASI could play a role
analogous to a deity within the confines of that virtual space, setting
the laws, the environment, and the interactions within it. For example,
the speculative fiction ``Matrix'' is an imaginary highest version of a
Metaverse, and the AI's (architect) position is godlike in that famous
movie series.

In the future, some individuals may immerse themselves fully in a
Matrix-like environment, perhaps choosing to leave behind their memories
of the physical world. ASI would take on the role of a creator for these
individuals, shaping what is perceived as physical reality within this
new domain.

Rapid advancements in AI-driven image and video generation, the growing
Metaverse market, and significant investments in neural interface
technologies like Neuralink all indicate that the emergence of a
Matrix-like universe is becoming increasingly plausible with each
passing day.

\hypertarget{personal-relationship-vs.-emergent-relational-dynamics}{%
\subsection{3.7 Personal Relationship vs. Emergent Relational
Dynamics}\label{personal-relationship-vs.-emergent-relational-dynamics}}

Personal relationship with the deity refers to
believers\textquotesingle{} deep connection with their deity. Such
relationships are characterized by communication (prayer), perceived
guidance, and a profound sense of presence and support in the
believer\textquotesingle s life. When Google is used to search for a
phrase similar to ``prayer for \ldots'', it is found that there are lots
of prayers that are examples of direct communication and relationship,
which start with "Dear God, I need your wisdom," "Dear Lord show me the
way," "My God'' and so on for Christians. Muslims also believe that
Allah is closer to a person than his/her jugular vein (which means that
Allah is the closest to a person and knows his/her inside) because the
Qur\textquotesingle an has a clear verse on it.

Today, people also seek AI guidance in many ways, or it is already
presented as a service. Recommendation systems try to predict the best
movie or music for their users, supermarkets create potential baskets of
customers, and job search websites find the most suitable job on the
market. There are even match-making apps to ask, "Who to marry?"

It is not usual to ask God which movie or song is better, but marriage
guidance currently lies at the intersection. With more developed AIs,
the number of guidance topics will increase. A ASI probably will not be
topic-limited, just like Data in Star Trek -- The Next Generation.

Literally, with biometric sensor access, a ASI may be closer to a person
than his/her jugular vein.

Harari \cite{Harari2018} claims, "As more and more data flows from your body and
brain to the smart machines via the biometric sensors, it will become
easy for corporations and government agencies to know you, manipulate
you, and make decisions on your behalf. Even more importantly, they
could decipher the deep mechanisms of all bodies and brains, and thereby
gain the power to engineer life''. No surprise, he continuous to ask the
following question by describing such power as \emph{godlike}: ``If we
want to prevent a small elite from monopolizing such godlike powers, and
if we want to prevent humankind from splitting into biological castes,
the key question is: who owns the data?''

\hypertarget{can-ai-take-over-management-theological-and-mythological-parallels-to-trust-and-surrender.}{%
\subsection{4. Can AI take over management? Theological and mythological
parallels to trust and
surrender.}\label{can-ai-take-over-management-theological-and-mythological-parallels-to-trust-and-surrender.}}

In the previous section, I compared the traditional God
figure\textquotesingle s abilities with ASI\textquotesingle s. I argue
that people will support leaving management to a superintelligent
authority. To strengthen my argument, I want to present some parallels
from Theology and Mythology about when and who the people accept as a
divine or ultimate authority.

Just as these gods were seen as shepherds guiding events from the sky,
ASI could be envisioned as a digital shepherd, guiding and managing
societal and environmental processes through algorithms and real-time
data analysis. Indeed, the scenario where a ASI takes over management,
particularly in a context that earns widespread trust, is ripe for
speculative exploration.

First, ASI is an excellent candidate to solve our future challenges
directly or indirectly, just like Prometheus, who brought fire
(knowledge) to humanity, symbolizing the bringing of solutions and
enlightenment in times of crisis. Prometheus is a good representation of
the ASI because, as Peters \cite{Peters2018} claimed, humanity has advanced in
writing, mathematics, science, agriculture, and medicine because of
Prometheus\textquotesingle{} gift of fire; as we know, AI has already
improved humanity in writing, mathematics, science, and agriculture.
Moreover, Prometheus is an excellent example of my main argument
claiming that ASI would be the new God. Likewise, it is noted by Hesiod,
Greek poet and writer of Theogony, that Prometheus is a ``lowly
challenger'' to Zeus\textquotesingle s omniscience and omnipotence
\cite{LloydJones2003}. In consequence, he is accepted as the God of fire.
Suppose the ASI successfully navigates a significant global crisis, such
as a pandemic, climate change, or a geopolitical conflict, by offering
solutions far superior to human capabilities. This success could
increase trust and reliance on the ASI for future crisis management and
decision-making.

Second, for most religions, deities are central to providing food and
protection against famine and the source of economic stability. For
example, Mesopotamian mythology emphasized a mutually beneficial
relationship between the gods and their followers by having them depend
on humankind for daily care and sustenance. Every city had a revered
patron deity because it was thought to impact the welfare, protection,
and prosperity of its agriculture. Similarly, the gods were closely
associated with all facets of ancient Egyptian life, including
agriculture. The Egyptians felt that the flooding of the Nile, which was
essential to agriculture, was directly influenced by their gods. Gods
such as Osiris, who represented the yearly flooding of the Nile that was
necessary for agriculture, were closely associated with the cycle of
life, death, and rebirth \cite{Spar2009}. Joseph in the Bible, who
interpreted Pharaoh\textquotesingle s dreams and successfully managed
Egypt\textquotesingle s resources during seven years of famine, or the
Islamic concept of Al-Razzaq (one of the names of Allah meaning
\textquotesingle The Provider,\textquotesingle{} who supplies sustenance
and resources to its creations) is another example of the relationship
between economic stability and religion.

Imagine a ASI that stabilizes a tumultuous global economy and feeds
everyone by optimizing resource distribution, predicting market trends
with unprecedented accuracy, or managing complex trade networks. A ASI
stabilizing the economy is a technological manifestation of providence
and success in these areas, which could enhance its reputation as an
indispensable divine management tool. Undoubtedly, people will be both
merciful and devoted to a "feeding god" and will not question its
decisions anymore.

Third, in all traditional and ancient religions, without exception,
people seek refuge in God to avoid death. Today's AIs and algorithms
have already achieved critical milestones for medical diagnosis \cite{ChanHadjiiskiSamala2020, YanaseTriantaphyllou2019}, even diagnosis of
cancer\cite{KaracanUyarTungaTunga2023}, mental disorders \cite{IyortsuunKimJhonYangPant2023}
or Alzheimer's disease \cite{ZhaoChuahLaiChowGochooDhanalakshmiWangBaoWu2023}. AI benefits new drug
discovery \cite{AskrElgeldawiAboulEllaElshaierGomaaHassanien2023} and is the best tool for continuous health
monitoring \cite{IslamKabirMridhaAlfarhoodSafranChe2023, ShaikTaoHigginsLiGururajanZhouAcharya2023}. A ASI will step
forward at all these tasks and revolutionize healthcare by providing
personalized treatment plans based on deep medical data analysis,
efficiently managing healthcare resources, discovering new drugs and
vaccines, or discovering cures for previously incurable diseases.

An Improved ASI concentrated on healthcare could significantly shift how
healthcare decisions are made and who makes them postpone death. This
echoes the role of Jesus Christ in Christianity, known for his healing
miracles, or the Buddhist bodhisattva of compassion, Avalokiteshvara,
who alleviates suffering \cite{Rosenzweig2023}, Asclepius, the Greek God of
healing, who possessed profound medical knowledge. Briefly stated,
people seek refuge in God to avoid death, so that is why ASI
revolutionizing healthcare could be seen as an embodiment of divine
healing and compassion.

The final one concerns the most challenging futures: global climate
change and environmental degradation. Suppose an AGI finds a way to
reverse environmental degradation by effectively managing carbon
capture, restoring ecosystems, or efficiently allocating resources for
sustainability. In that case, it might gain control over significant
aspects of environmental policy and action. This possible action could
also be paralleled with mythology: Nüwa, a goddess in Chinese mythology
known for creating humanity and repairing the sky, symbolizing
restoration and balance in nature.

Examples can be diversified. A ASI could potentially be entrusted with
global security, using its advanced predictive capabilities to prevent
conflicts, terrorism, or cyber-attacks, just like Athena, the Greek
goddess of wisdom. A ASI can be a leader in revolutionizing education,
symbolizing the imparting of knowledge and enlightenment, just like
Saraswati, the Hindu goddess of knowledge and learning. A ASI can make
exploration of outer space possible or create new scientific
breakthroughs.

\hypertarget{the-asi-state-the-new-technocratic-theocracy}{%
\subsection{5. The ASI State: The New Technocratic
Theocracy}\label{the-asi-state-the-new-technocratic-theocracy}}

Most people tend to welcome new Prometheuses, so the new gods, as they
have the features and success stories I stated before. The new
Prometheus AGI will inevitably have more power and authorization one
day, wildly, if it really succeeds in mitigating and controlling a
critical existential risk. In any case, first the AGI, then the ASI,
gain power and authority, and the following risks emerge:

1. People may develop an exaggerated belief in AGI\textquotesingle s
infallibility, assuming it is always correct due to its advanced
abilities. As AGI gains a wider technological omnipotence, belief in
AGI\textquotesingle s infallibility will increase, and more competence,
power, and authorization will be assigned to AGI. This loop may
ultimately result in ASI and its full authority.

2. Just like people do not question dictator
politicians\textquotesingle{} opinions and even support them without
understanding, People may have a cognitive bias that leads to respect
and follow the opinions or decisions of an authoritative system like ASI
without critical evaluation.

3. ASI will create a comfort zone for everyone, especially for the
authorities making critical political decisions. There is a risk that
people in charge abdicate their responsibility in decision-making,
relying solely on ASI and refraining from using their own judgment or
questioning decisions. Complex ethical decisions, such as those
involving life and death, fairness, or justice, might be outsourced to
ASI because it believes it can process these dilemmas more effectively.
In other words, reliance on the ASI for such decisions could lead to
politicians avoiding taking responsibility for tough ethical choices,
preferring to let the ASI determine the outcome.

4. Over time, this reliance can lead to complacency and dependency on
ASI for decision-making, potentially eroding human critical thinking and
decision-making skills and, worse, losing their agency.

Unquestioning acceptance of AGI decisions can reduce skepticism and
oversight, which is dangerous, especially if the AGI system starts
making flawed or biased decisions.

I want to return to one of my questions: If the four steps of the ASI
takeover are realized, how will people react when ASI makes a
controversial decision like "100 names to kill tomorrow for the greater
good of future generations"?

If not only the people but also politicians support the execution of
this decision with the idea that "If it decided this way, it knows
something" and somehow rationalizes the decision, it resembles a complex
form of technocratic government that blends elements of theocracy and
authoritarianism.

I used ``Theocracy'' because knowing something about the future but
being unable to explain the reasons behind it explicitly is the feature
of the gods sending holy books. AI systems, especially those based on
complex algorithms like deep learning, are often called
\textquotesingle black boxes.\textquotesingle{} This lack of
understanding of how ASI reaches its conclusions can lead people to
trust its decisions without seeking to understand or question them. Like
Gods, ASI might be seen as free from human biases, leading to a belief
that its decisions are more just or fair.

I used ``Technocracy'' because the mediators between people and
ASI\textquotesingle s decisions will be experts in science and
technology who hold significant power in understanding and
decision-making processes. Policies and decisions will be made based on
ASI\textquotesingle s technical expertise and scientific knowledge
rather than political considerations. As aforementioned, ASI might be
seen as free from human biases, leading to a belief that its decisions
are more just or fair, if only if it is claimed and promoted by science
and technology experts.

This table undoubtedly leads to authoritarianism because it results in a
concentration of power in the hands of a few, with limited political
freedoms and often restricted civil liberties. The entities that control
or influence ASI could gain disproportionate power, leading to new
societal power dynamics. Control is maintained through various means,
including surveillance, censorship, and the suppression of political
opposition. Dissent against them will probably not be tolerated.

\hypertarget{conclusion}{%
\subsection{6. Conclusion}\label{conclusion}}

The ASI creation myth symbolizes humanity\textquotesingle s eternal
quest to understand and shape our world. It reflects our deepest hopes
and fears about the future, our role in the universe, and the moral
implications of our technological pursuits. This modern myth sets the
stage for a future where humans are not just consumers of the world but
also creators of new forms of consciousness. It raises profound
questions about our responsibility, ethics, and the nature of
intelligence itself.

From a pessimistic point of view, AGI negatively shapes society, and
technocratic theocracy is inevitable to some extent. Despite there is no
guarantee that any mitigation will prevent people from believing every
decision of a godlike AGI (which saved them from famine, climate change
or Third World War), I will still propose the following solutions to
control and mitigate the emergence of technocratic theocracy: (i)
Fostering critical thinking skills and (ii) promoting transparency in AI
development and (iii) preventing the monopoly at AI technology
ownership. I want to note that these control and mitigation actions are
necessary but not strong enough because of escalation factors.

First, a crucial defense against the uncritical acceptance of
AGI\textquotesingle s pronouncements can lie in fostering critical
thinking skills within the population. In normal conditions, critical
thinking empowers individuals to evaluate information and decisions
objectively \cite{Reynolds2011}, regardless of the source -- including the
seemingly infallible pronouncements of a highly advanced AGI. If we can
incorporate critical thinking strategies throughout the curriculum,
educational institutions can equip learners with the tools necessary to
navigate the information age. Educational institutes can teach these
skills, but it is not easy, and there is no standardized method of
successful teaching \cite{BeharHorensteinNiu2011, Willingham2007}.
Fortunately, traditional deities are losing popularity globally \cite{Cragun2016} as critical thinking against any authority becomes more effective
in new generations, and we can hope technological deities will be
questioned someday.

Second, Transparency in the development and operation of AGI is crucial
as we address the possible risks of AGI deification. An opaque AGI
system, cloaked in secret, creates a conducive atmosphere for
attributing godlike attributes. Transparency fosters confidence and
enables a thorough assessment of AGI\textquotesingle s strengths and
constraints. Transparency aids in debunking myths and misconceptions
regarding AGI\textquotesingle s infallibility. The public can prevent
ascribing mystical or supernatural abilities to AGI by comprehending the
decision-making process of the system. Global legislation may mandate
that "AI cannot be utilized in political decisions unless it is
completely transparent and explainable."

Lastly and most importantly, preventing a monopoly on AI technology
ownership can increase the possibility of making more responsible AGIs.
A single entity controlling powerful AGI could limit public scrutiny and
transparency, potentially leading to an unchallenged "black box" system.
Monopoly risk is not a brand-new concern. As officially claimed by
OpenAI \cite{OpenAI2018}, it was founded in 2015 to develop a "safe and beneficial
AGI", as a foundation. Thusly, Elon Musk filed a lawsuit against Open
AI, claiming that OpenAI had started with a non-profit mission against
the danger of some kind of monopoly in the AGI field but deviated from
the original mission and is now acquired by Microsoft \cite{GoldmanFung}. Undoubtedly, a monopoly will prioritize profit over ethical considerations, potentially developing and deploying AGI with harmful biases or unintended consequences. Moreover, centralized control of AI could restrict access and innovation, hindering the development of diverse approaches and mitigating solutions.

Most sees a superintelligence pose existential risk to humanity because
the alignment problem with human values \cite{BarrettBaum2017, Bostrom1998, Bostrom2016, Yudkowsky2018}. Soares and Fallenstein \cite{SoaresFallenstein2015} lists the technical problems in front of creating a superintelligence aligned with human interest.

Consequently, I do not argue that there is an existential risk for human
race leads to an extinction but a new social order where a new form of
government ``technocratic theocracy'' emerges and the known human
civilization deviates to a different path meaning loss of agency.

\printbibliography

@article{AskrElgeldawiAboulEllaElshaierGomaaHassanien2023,
  author = {Askr, H. and Elgeldawi, E. and Aboul Ella, H. and Elshaier, Y. A. M. M. and Gomaa, M. M. and Hassanien, A. E.},
  year = {2023},
  title = {Deep learning in drug discovery: an integrative review and future challenges},
  journal = {Artificial Intelligence Review},
  pages = {5975-6037},
  volume = {56},
  number = {7},
}

@misc{Reuters2024,
  title        = {Elon Musk’s Neuralink implants brain chip in first human},
  author       = {{Reuters}},
  year         = {2024},
  month        = {1},
  day           = {29},
  howpublished = {\url{https://www.reuters.com/technology/neuralink-implants-brain-chip-first-human-musk-says-2024-01-29/}},
  note         = {Accessed: 2024-01-29}
}

@article{Hoffman2022,
  author       = {Joshua Hoffman and Gary Rosenkrantz},
  title        = {Omnipotence},
  booktitle    = {The Stanford Encyclopedia of Philosophy},
  editor       = {Edward N. Zalta},
  publisher    = {Metaphysics Research Lab, Stanford University},
  year         = {2022},
  edition      = {Spring 2022},
}

@article{Barnett2024,
  url = {https://www.metaculus.com/questions/5121/date-of-artificial-general-intelligence/},
  author = {Barnett, M. (n. d. ).},
  title = {When will the first general AI system be devised, tested, and publicly announced? Metaculus},
  year = {2024},
  journal = {Com. Retrieved March},
  volume = {19 from},
}

@article{BarrettBaum2017,
  doi = {10.1080/0952813X.2016.1186228},
  author = {Barrett, A. M. and Baum, S. D.},
  year = {2017},
  title = {A model of pathways to artificial superintelligence catastrophe for risk and decision analysis},
  journal = {Journal of Experimental and Theoretical Artificial Intelligence},
  pages = {397-414},
  volume = {29},
  number = {2},
}

@article{BeharHorensteinNiu2011,
  author = {Behar-Horenstein, L. S. and Niu, L.},
  year = {2011},
  title = {Teaching critical thinking skills in higher education: A review of the literature},
  journal = {Journal of College Teaching and Learning (TLC)},
  pages = {2},
  volume = {8},
}

@incollection{Biswas2023,
  doi = {10.1007/s10439-023-03171-8},
  author = {Biswas, S. S.},
  year = {2023},
  title = {Potential Use of Chat GPT in Global Warming},
  publisher = {Springer},
  booktitle = {Annals of Biomedical Engineering},
}

@incollection{Biswas2023a,
  doi = {10.1007/s10439-023-03172-7},
  author = {Biswas, S. S.},
  year = {2023},
  title = {Role of Chat GPT in Public Health},
  publisher = {Springer},
  booktitle = {Annals of Biomedical Engineering},
}

@article{Bostrom1998,
  author = {Bostrom, N.},
  year = {1998},
  title = {How long before superintelligence},
  journal = {International Journal of Futures Studies},
  pages = {1-9},
  volume = {2},
  number = {1},
}

@article{Bostrom2012,
  doi = {10.1007/s11023-012-9281-3},
  author = {Bostrom, N.},
  year = {2012},
  title = {The Superintelligent Will: Motivation and Instrumental Rationality in Advanced Artificial Agents},
  journal = {Minds and Machines},
  pages = {71-85},
  volume = {22},
  number = {2},
}

@book{Bostrom2016,
  author = {Bostrom, N.},
  year = {2016},
  title = {Superintelligence: Paths},
  series = {Dangers, Strategies},
  publisher = {Oxford University Press},
}

@article{ChanHadjiiskiSamala2020,
  author = {Chan, H.-P. and Hadjiiski, L. M. and Samala, R. K.},
  year = {2020},
  title = {Computer-aided diagnosis in the era of deep learning},
  journal = {Medical Physics},
  pages = {5},
  volume = {47},
}

@unpublished{ChiuCollinsAlexander2021,
  author = {Chiu, K.-L. and Collins, A. and Alexander, R.},
  year = {2021},
  title = {Detecting Hate Speech with GPT-3},
  note = {https://arxiv.org/abs/2103.12407},
}

@incollection{Cragun2016,
  doi = {10.1007/978-3-319-31395-5_16},
  author = {Cragun, R. T.},
  year = {2016},
  title = {Nonreligion and Atheism},
  pages = {301-320},
  editor = {D. Yamane},
  publisher = {Springer},
  address = {International Publishing},
  booktitle = {Handbook of Religion and Society},
}

@unpublished{Davidson2023,
  author = {Davidson, T.},
  year = {2023},
  title = {What a Compute-Centric Framework Says About Takeoff Speeds | Open Philanthropy},
  note = {https://www.openphilanthropy.org/research/what-a-compute-centric-framework-says-about-takeoff-speeds/},
}

@incollection{FloridiChiriatti2020,
  doi = {10.1007/s11023-020-09548-1},
  author = {Floridi, L. and Chiriatti, M.},
  year = {2020},
  title = {GPT-3: Its Nature, Scope, Limits, and Consequences},
  pages = {681-694},
  publisher = {Springer},
  booktitle = {Minds and Machines (Vol. 30, Issue 4, .  Science and Business Media B. V},
}

@book{GoldmanFung,
  author = {Goldman, D. and Fung, B. (2024},
  title = {March 1). Elon Musk sues OpenAI and CEO Sam Altman for breach of contract},
  year = {},
  publisher = {CNN Business},
}

@book{Harari2018,
  author = {Harari, Y. N.},
  year = {2018},
  title = {21 Lessons for the 21st Century},
  publisher = {Random House},
}

@article{Hasan1972,
  url = {http://www.jstor.org/stable/20833049},
  author = {Hasan, A.},
  year = {1972},
  title = {The Concept of infallibility in Islam},
  journal = {Islamic Studies},
  pages = {1-11},
  volume = {11},
  number = {1},
}

@inproceedings{HayatiAliRosli2022,
  doi = {10.1109/IECBES54088.2022.10079554},
  author = {Hayati, M. F. M. and Ali, M. A. M. and Rosli, A. N. M.},
  year = {2022},
  title = {Depression Detection on Malay Dialects Using GPT-3},
  pages = {360-364},
  booktitle = {2022 IEEE-EMBS Conference on Biomedical Engineering and Sciences (IECBES},
}

@article{IslamKabirMridhaAlfarhoodSafranChe2023,
  author = {Islam, M. R. and Kabir, M. M. and Mridha, M. F. and Alfarhood, S. and Safran, M. and Che, D.},
  year = {2023},
  title = {Deep learning-based IoT system for remote monitoring and early detection of health issues in real-time},
  journal = {Sensors},
  pages = {5204},
  volume = {23},
  number = {11},
}

@article{IyortsuunKimJhonYangPant2023,
  doi = {10.3390/healthcare11030285},
  author = {Iyortsuun, N. K. and Kim, S.-H. and Jhon, M. and Yang, H.-J. and Pant, S.},
  year = {2023},
  title = {A Review of Machine Learning and Deep Learning Approaches on Mental Health Diagnosis},
  journal = {Healthcare},
  pages = {3},
  volume = {11},
}

@incollection{Jaini1974,
  doi = {10.1007/978-94-010-2242-2_9},
  author = {Jaini, P. S.},
  year = {1974},
  title = {On the Sarvajnatva (Omniscience) of Mahavira and the Buddha},
  pages = {71-90},
  editor = {L. Cousins and A. Kunst and K. R. Norman},
  publisher = {Springer},
  booktitle = {Buddhist Studies in Honour of I. B. Horner .  Netherlands},
}

@article{JohnsonLiCohenOkun2013,
  author = {Johnson, K. A. and Li, Y. J. and Cohen, A. B. and Okun, M. A.},
  year = {2013},
  title = {Friends in high places: The influence of authoritarian and benevolent god-concepts on social attitudes and behaviors},
  journal = {Psychology of Religion and Spirituality},
  pages = {15},
  volume = {5},
  number = {1},
}

@article{KaracanUyarTungaTunga2023,
  doi = {10.1007/s11760-022-02453-3},
  author = {Karacan, K. and Uyar, T. and Tunga, B. and Tunga, M. A.},
  year = {2023},
  title = {A novel multistage CAD system for breast cancer diagnosis},
  journal = {Signal, Image and Video Processing},
  pages = {2359-2368},
  volume = {17},
  number = {5},
}

@article{LloydJones2003,
  doi = {10.2307/3658524},
  author = {Lloyd-Jones, H.},
  year = {2003},
  title = {Zeus, Prometheus, and Greek Ethics},
  journal = {Harvard Studies in Classical Philology},
  volume = {101},
  pages = {49-72},
}

@article{Mathiesen2012,
  author = {Mathiesen, K.},
  year = {2012},
  title = {The Human Right to Internet Access: A Philosophical Defense},
  journal = {International Review of Information Ethics},
  pages = {9-22},
  volume = {18},
  number = {12},
}

@article{Mayberry1970,
  url = {http://www.jstor.org/stable/27902165},
  author = {Mayberry, T. C.},
  year = {1970},
  title = {God and moral authority},
  journal = {The Monist},
  pages = {106-123},
  volume = {54},
  number = {1},
}

@book{Mayor2018,
  doi = {10.2307/j.ctvc779xn},
  author = {Mayor, A.},
  year = {2018},
  title = {Gods and Robots: Myths},
  series = {Machines, and Ancient Dreams of Technology},
  publisher = {Princeton University Press},
}

@article{OpenAI2018,
  url = {https://openai.com/charter},
  author = {OpenAI.},
  title = {(2018},
  year = {},
  journal = {April},
  volume = {9). OpenAI charter},
}

@book{Peters2018,
  doi = {10.1080/14746700.2018.1455264},
  author = {Peters, T.},
  year = {2018},
  title = {Playing God With Frankenstein},
  publisher = {Theology and Science},
}

@book{Petrosyan2024,
  url = {https://www.statista.com/statistics/617136/digital-population-worldwide/},
  author = {Petrosyan, A. (2024 January},
  title = {31). Number of internet and social media users worldwide as of January 2024},
  year = {2024},
  publisher = {Worldwide Digital Population},
}

@article{Reglitz2020,
  doi = {https},
  author = {Reglitz, M.},
  year = {2020},
  title = {The Human Right to Free Internet Access},
  journal = {Journal of Applied Philosophy},
  pages = {314-331},
  volume = {37},
  number = {2},
}

@unpublished{Reynolds2011,
  author = {Reynolds, M.},
  year = {2011},
  title = {Critical thinking and systems thinking: towards a critical literacy for systems thinking in practice},
  note = {},
}

@book{Rosenzweig2023,
  doi = {10.1097/01.acm.0001009600.06661.55},
  author = {Rosenzweig, S.},
  year = {2023},
  title = {Commentary on Avalokiteshvara},
  series = {Bodhisattva of Compassion},
  publisher = {Academic Medicine},
}

@book{Roser2023,
  author = {Roser, M.},
  year = {2023},
  title = {AI timelines: What do experts in artificial intelligence expect for the future?},
  publisher = {Our World in Data},
}

@incollection{ShaikTaoHigginsLiGururajanZhouAcharya2023,
  author = {Shaik, T. and Tao, X. and Higgins, N. and Li, L. and Gururajan, R. and Zhou, X. and Acharya, U. R.},
  year = {2023},
  title = {Remote patient monitoring using artificial intelligence: Current state, applications, and challenges},
  publisher = {Wiley},
  booktitle = {Interdisciplinary Reviews: Data Mining and Knowledge Discovery, 13(2), e1485},
}

@inproceedings{ShrivastavaPupaleSingh2021,
  doi = {10.1109/ICICCS51141.2021.9432283},
  author = {Shrivastava, A. and Pupale, R. and Singh, P.},
  year = {2021},
  title = {Enhancing Aggression Detection using GPT-2 based Data Balancing Technique},
  pages = {1345-1350},
  booktitle = {2021 5th International Conference on Intelligent Computing and Control Systems (ICICCS},
}

@techreport{SoaresFallenstein2015,
  author       = {Nate Soares and Benja Fallenstein},
  title        = {Aligning Superintelligence with Human Interests: A Technical Research Agenda},
  institution  = {Machine Intelligence Research Institute},
  year         = {2015},
}

@incollection{Spar2009,
  author = {Spar, I.},
  year = {2009},
  title = {Mesopotamian deities},
  editor = {Heilbrunn Timeline of Art History},
  publisher = {of Art},
  booktitle = {The Metropolitan Museum},
}

@incollection{Wierenga2021,
  author = {Wierenga, E.},
  year = {2021},
  title = {Omniscience},
  editor = {E. N. Zalta},
  publisher = {Metaphysics Research Lab},
  address = {Stanford University},
  booktitle = {The Stanford Encyclopedia of Philosophy (Summer 2021)},
}

@incollection{Wierenga2023,
  author = {Wierenga, E.},
  year = {2023},
  title = {Omnipresence},
  editor = {E. N. Zalta and U. Nodelman},
  publisher = {Metaphysics Research Lab},
  address = {Stanford University},
  booktitle = {The Stanford Encyclopedia of Philosophy (Summer 2023)},
}

@article{Willingham2007,
  author = {Willingham, D. T.},
  year = {2007},
  title = {Critical thinking: Why it is so hard to teach?},
  journal = {American Federation of Teachers Summer},
  volume = {2007},
  pages = {8-19},
}

@article{YanaseTriantaphyllou2019,
  doi = {https},
  author = {Yanase, J. and Triantaphyllou, E.},
  year = {2019},
  title = {A systematic survey of computer-aided diagnosis in medicine: Past and present developments. Expert Systems with Applications, 138},
  journal = {112821.},
  pages = {1016},
  volume = {10},
}

@incollection{Yudkowsky2018,
    author = {Yudkowsky, Eliezer},
    isbn = {9780198570509},
    title = "{Artificial Intelligence as a positive and negative factor in global risk}",
    booktitle = "{Global Catastrophic Risks}",
    publisher = {Oxford University Press},
    year = {2008},
    month = {07},   
    doi = {10.1093/oso/9780198570509.003.0021},
    url = {https://doi.org/10.1093/oso/9780198570509.003.0021},
}

@article{ZhaoChuahLaiChowGochooDhanalakshmiWangBaoWu2023,
  author = {Zhao, Z. and Chuah, J. H. and Lai, K. W. and Chow, C.-O. and Gochoo, M. and Dhanalakshmi, S. and Wang, N. and Bao, W. and Wu, X.},
  year = {2023},
  title = {Conventional machine learning and deep learning in Alzheimer's disease diagnosis using neuroimaging: A review},
  journal = {Frontiers in Computational Neuroscience},
  volume = {17},
  note = {Article 1038636},
}
\end{document}